\documentclass[letterpaper, 10 pt, conference]{ieeeconf}  

\IEEEoverridecommandlockouts                              

\usepackage{times}
\usepackage[pdftex]{graphicx}
\usepackage{subfigure}
\usepackage{amsmath,amssymb,amsopn,amstext,amsfonts}
\usepackage{algorithm,algorithmic}
\usepackage{cancel}
\usepackage[space]{cite}
\usepackage{pdfsync}
\usepackage{balance}
\usepackage{color}
\usepackage{mathtools}
\usepackage{bm}

\usepackage{diagbox}
\usepackage{float}
\usepackage{epstopdf}
\usepackage{pifont}
\usepackage{fixltx2e} 
\usepackage{amsmath}
\usepackage{multirow}
\usepackage{url}
\usepackage{verbatim}
\usepackage[linkcolor=black,citecolor=black,urlcolor=black,colorlinks=true]{hyperref}
\usepackage{soul,xcolor}
\usepackage{subfiles}
\graphicspath{../figure}
\DeclareGraphicsExtensions{.png,.jpg,.eps,.pdf,.jpeg}

\usepackage{ulem}

\title{\LARGE \bf
Exploration with Global Consistency \\ Using Real-Time Re-integration and Active Loop Closure}

\author{Yichen Zhang, Boyu Zhou, Luqi Wang and Shaojie Shen%
\thanks{This work was supported by HKUST Postgraduate Studentship and HDJI Lab. All authors are with the Department of Electronic and Computer Engineering, Hong Kong University of Science and Technology, Hong Kong, China. {\tt\footnotesize $\{$yzhangec, bzhouai, lwangax, eeshaojie$\}$@connect.ust.hk}}%
}

\begin{document}
\maketitle
\thispagestyle{empty}
\pagestyle{empty}
 
\begin{abstract}
  Despite recent progress of robotic exploration, most methods assume that drift-free localization is available, which is problematic in reality and causes severe distortion of the reconstructed map. In this work, we present a systematic exploration mapping and planning framework that deals with drifted localization, allowing efficient and globally consistent reconstruction. A real-time re-integration-based mapping approach along with a frame pruning mechanism is proposed, which rectifies map distortion effectively when drifted localization is corrected upon detecting loop-closure. Besides, an exploration planning method considering historical viewpoints is presented to enable active loop closing, which promotes a higher opportunity to correct localization errors and further improves the mapping quality. We evaluate both the mapping and planning methods as well as the entire system comprehensively in simulation and real-world experiments, showing their effectiveness in practice. The implementation of the proposed method will be made open-source for the benefit of the robotics community.
\end{abstract}


\section{Introduction}
\label{sec:intro}


\begin{figure}[t] 
  \begin{center}
    {\includegraphics[width=0.7\columnwidth]{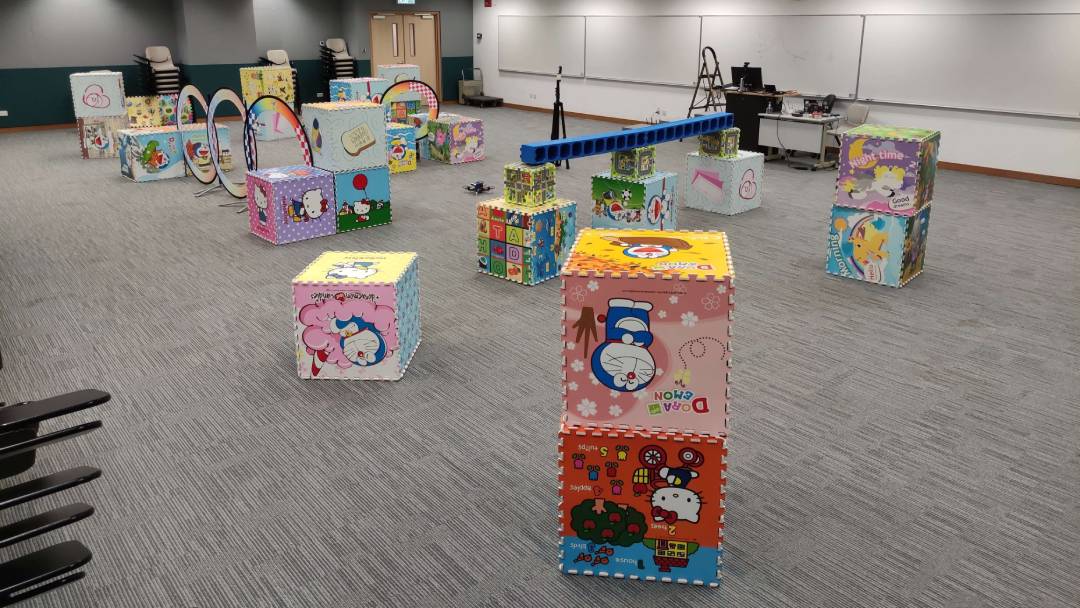} \label{fig:scene}}
    {\includegraphics[width=0.7\columnwidth]{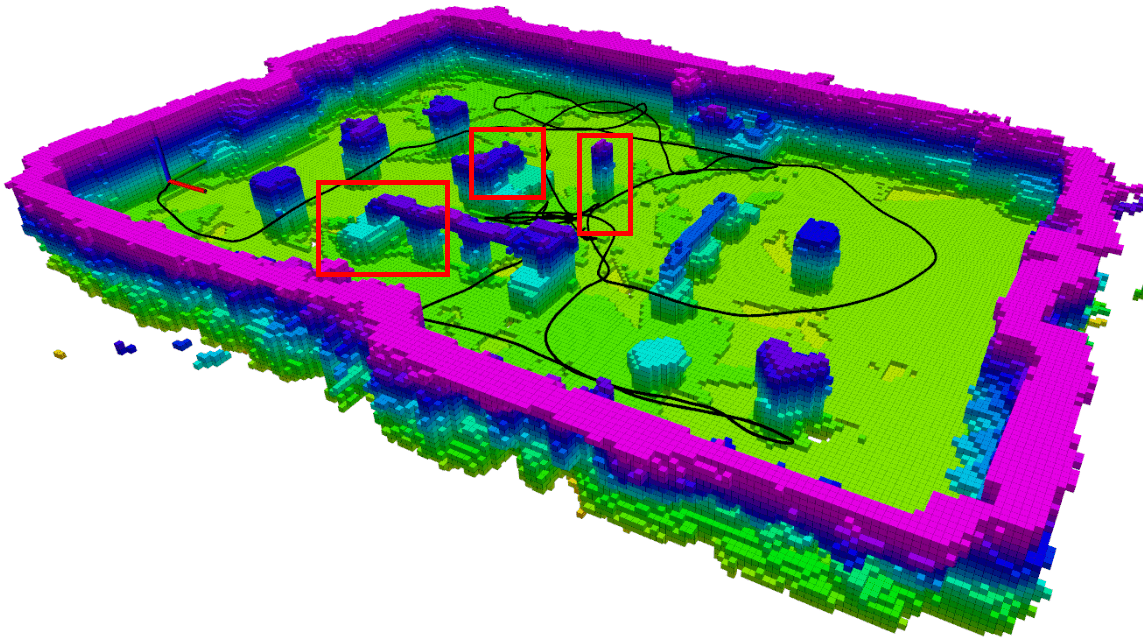} \label{fig:all2}}
    {\includegraphics[width=0.7\columnwidth]{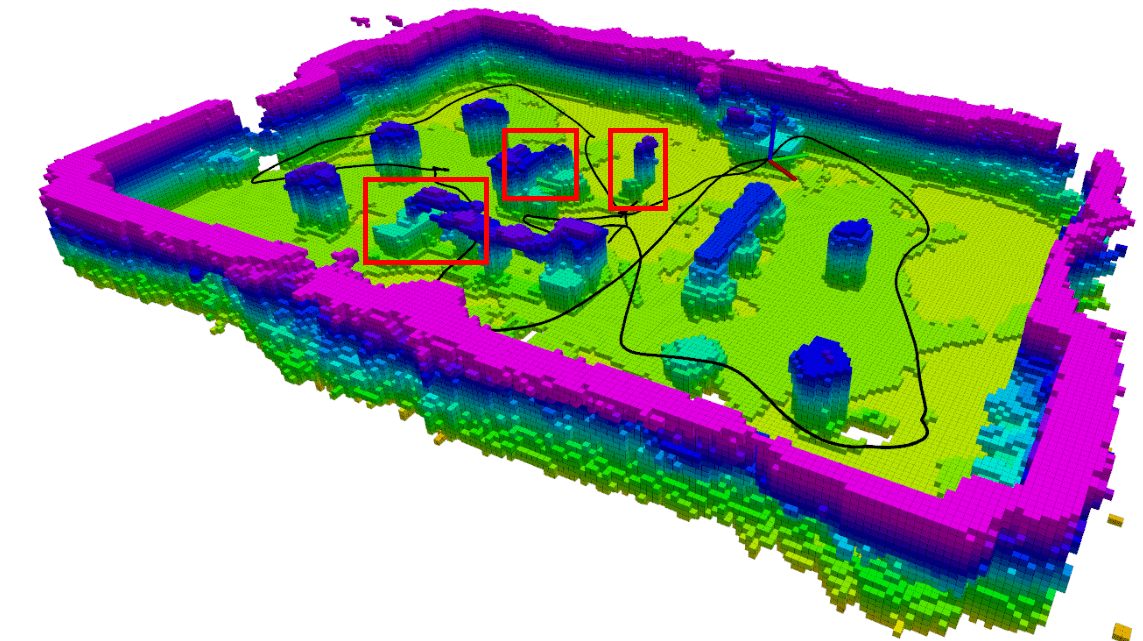} \label{fig:all1}}
    \vspace{-0.4cm}
  \end{center}
  \caption{\label{fig:intro} The quadrotor autonomous exploration tests are conducted in a complex indoor scene with size \(20 \times 14 \times 3 \text{m}^3\). From top to bottom: Actual scene of experiment field; Our exploration results; FUEL\cite{zhou2021fuel} results. Significant differences are marked with red box.}
  \vspace{-0.8cm}
\end{figure}

Autonomous exploration, which sends robots to completely map an unknown environment, has been a popular topic in robotics research,
because it is a fundamental problem for a variety of applications such as aerial photography, geographic mapping, disaster management, and search and rescue.
During the past decades, many algorithms have been proposed to improve exploration performance \cite{zhou2021fuel, shen2012stochastic, keidar2014efficient}.

Despite recent progresses, most of the current methods concern only about the exploration efficiency and give less consideration about the map quality. 
In particular, many of them assume that perfect localization is available throughout the exploration, so accurate reconstruction is easily obtainable as well.
Unfortunately, in practice, accumulating drift of onboard state estimation is inevitable in a long run.
The drift often has a negative impact on the reconstructed map, such as causing significant distortions.
Consequently, the resulting map may be of poor quality, which is unsatisfactory for many applications that require accurate reconstruction.
Therefore, to enable a broader range of applications, the mapping algorithm should be able to rectify the distortion caused by localization drift explicitly.
Moreover, previous methods seldom account for potential opportunities to detect loops, although the map quality is highly dependent on the detection of loop closure.
Instead, detecting loop closure during exploration is mostly passive and depends on whether the robot happens to pass a previously visited place.
Hence, the exploration algorithm should seek for more opportunities on loop closing without sacrificing much efficiency to improve map quality further.


In order to solve the two issues mentioned above, we propose a systematic solution for autonomous exploration, which enables the generation of accurate and globally consistent maps. 
First, we introduce a TSDF-based mapping framework with real-time re-integration and a frame pruning mechanism based on a set cover formulated selection criterion. 
When loop closure is detected and drift is eliminated by the localization module, the map distortion caused by localization drift is rectified by re-integrating the keyframes with updated camera poses, resulting in a globally consistent map. 
To enable real-time re-integration on the onboard computer, we novelly apply a keyframe selection method based on set cover formulation to prune the redundant frames. Then, we utilize space partitioning to incrementally solve the set cover problem, increasing the real-time performance.
Moreover, we adopt an active loop closing strategy in exploration planning to allow a higher opportunity to detect loops, which further reduces localization drift and enhances map quality.



Extensive experiments are conducted in both benchmarked simulations and the real world to validate the performance of our proposed method. In the simulation, we compare our proposed framework with the results from other mapping approaches. The results show that our proposed method outperforms other works, especially under severe odometry drift. 
In real world tests, our proposed method generates a more globally consistent exploration result, comparing with the state-of-the-art exploration system \cite{zhou2021fuel}.
The contributions of this work can be summarized in the following:
\begin{enumerate}
  \item A globally consistent mapping framework with online re-integration and frame pruning mechanism, which can rectify the map distortion.
  \item An active loop closing planning strategy which allows more loops to be detected thus further improves the mapping quality.
  \item Benchmark comparison experiments and field tests are conducted to evaluate the performance of our proposed method. The source code of our work will be published.
\end{enumerate}


\section{Related Work}
\label{sec:related}

\subsection{Autonomous Exploration}
\label{subs:related_exploration}






Robotic exploration has been investigated over the years.
Some works aim at quick coverage\cite{cieslewski2017rapid, dharmadhikari2020motion}, while others emphasize precise mapping\cite{schmid2020efficient, song2017online}.
Frontier-based approaches are presented earliest in \cite{yamauchi1997frontier} and assessed more comprehensively later in \cite{julia2012comparison}.
In \cite{shen2012stochastic} a stochastic differential equation-based strategy is used for 3D environments.
Unlike\cite{yamauchi1997frontier} choosing the closest frontier as the next target, \cite{cieslewski2017rapid} chooses the frontier inside the FOV that minimizes the velocity change.
The strategy is advantageous for a high flight speed and shows higher efficiency than \cite{yamauchi1997frontier}.
\cite{deng2020robotic} presented a differentiable formulation of information gain, which is useful for gradient-based optimization of the exploration path.

Another major category is sampling-based approaches, which produce viewpoints randomly to explore the space.
These approaches are closely related to next best view (NBV) \cite{connolly1985determination}, where views are computed repeatedly to model a scene completely.
\cite{bircher2016receding} grows RRTs and executes the most informative edge repeatedly.
Uncertainty \cite{papachristos2017uncertainty}, visual importance \cite{dang2018visual},inspection \cite{bircher2018receding} and history\cite{witting2018history, wang2019efficient} are considered later under this framework\cite{bircher2016receding}.

Approaches combining the advantages of frontier-based and sampling-based approaches are presented in \cite{charrow2015information,selin2019efficient,meng2017two,song2017online,caoexploring,yang2021graph, zhou2021fuel}.
\cite{charrow2015information,selin2019efficient} generate global paths based on frontiers and sample paths locally.
A two-level framework \cite{caoexploring} computes exploration paths coarsely at the global scale and finely around the robot.
In \cite{yang2021graph}, sparse topological graphs are built to provide high-level guidance.
\cite{zhou2021fuel} proposed a hierarchical planning framework based on the incrementally extracted frontiers, which shows high-performance exploration in complex environments.

A common issue of existing methods is that they seldom consider the rectification of map under odometry drift, except \cite{schmid2021glocal}, which may not produce globally consistent map in practice. In this work, we deal with this problem systematically.

\subsection{Globally Consistent Dense Mapping}
\label{subs:related_reconstruction}

Researchers have suggested a number of approaches to rectify distorted map when loop-closures are detected.
One idea is to attach a deformable mesh \cite{whelan2012kintinuous} to a pose graph representing the past trajectory of the camera.
Alternatively, a deformable collection of surfels may be maintained \cite{whelan2015elasticfusion}.
Although these approaches provide a richer understanding of the occupied space, the free space and unknown space are not distinguished, so they are not suitable for robotic exploration where information of unknown space is essential.

Another approach is building a collection of submaps to present the global map. The relative transformation of each submap is corrected when localization is rectified by loop-closure, which can maintain map consistency efficiency.
Several works have shown the efficacy of such approaches, including \cite{millane2018c, reijgwart2019voxgraph, wagner2014graph, fioraio2015large}
One limitation of these methods is that they put the burden on the motion planner, which has to query map information from multiple submaps.
Besides, they can not eliminate distortion completely under severe localization drift.

\cite{dai2017bundlefusion} proposes a re-integration method to maintain a globally consistent map, which stores sensor measurements and associated sensor poses to rectify the map upon localization correction.
It produces higher reconstruction quality, but requires significant computational resources.
\cite{han2018flashfusion} improves the computational efficiency by a large margin. However, it does not explicitly represent unknown space.
In contrast to submaps, these works provide constant query time for motion planning module.
In this work, a mapping method based on re-integration is adopted and further optimized by a frame pruning technique, enabling real-time mapping in exploration.


\begin{figure}[t]
  \begin{center}
    {\includegraphics[width=0.8\columnwidth]{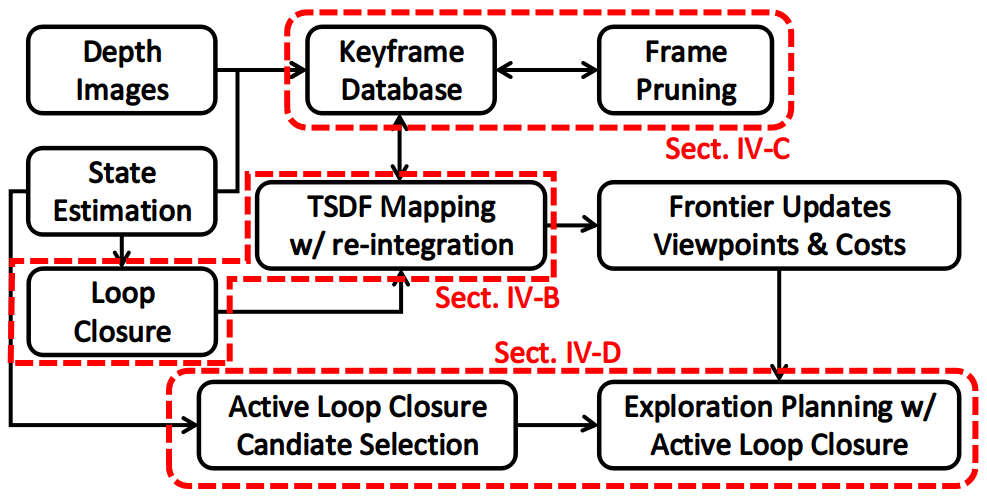}}
  \end{center}
  \vspace{-0.4cm}
  \caption{System overview of our proposed exploration framework.}
  \label{fig:overview}
  \vspace{-1.0cm}
\end{figure}

\section{System Overview}
\label{sec:overview}

An overview of our proposed systematic solution for autonomous exploration can be found in Fig. \ref{fig:overview}. With the depth measurement and onboard state estimation, the mapping module can carry out a globally consistent reconstruction for exploration planning. The re-integration (Sect. \ref{method: reintegration}) of rectified point cloud plays a key role in eliminating map distortion caused by pose drift during the flight. We use a frame pruning method (Sect. \ref{method: keyframe}) formulated as a set cover problem to enable online re-integration. Besides considering frontiers in space, possible loop closure viewpoints (Sect. \ref{method:alc}) are also included in autonomous exploration planning for better global consistency.

\section{Methodology}
\subsection{TSDF-based 3D Mapping}
\label{method: map}
The mapping module of the proposed method is based on Truncated Signed Distance Field (TSDF), whose construction method is similar to \cite{oleynikova2017voxblox,dai2017bundlefusion, newcombe2011kinectfusion}. The key difference is that our framework allows real-time and globally consistent reconstruction by adopting the re-integration (Sect.\ref{method: reintegration}) and frame pruning (Sect.\ref{method: keyframe}) methods.

We take the point cloud generated from the depth image as an input frame.
For each frame, ray-casting is performed from camera position \(^w C\) to each point \(^w P\) in the point cloud and all voxels along the ray are updated with new distance value \(sdf\) and weight \(w\). The distance value \(sdf\) can be computed as the length difference of the vector \(\overrightarrow{^wC^wP}\) and \(\overrightarrow{^wC^wV}\), where \(^wV\) is the point that the center of voxel \(v\) projected on the ray \(\overrightarrow{^wC^wP}\). Note that the \(sdf\) value is bounded to range \([-\delta, \delta]\) as \cite{dai2017bundlefusion} to support de-integration process, where \(\delta = t\sigma_{map}\) is the truncated distance and \(t\) is a scaling factor and \(\sigma_{map}\) is the map resolution.

\vspace{-0.4cm}
\begin{equation}
  sdf = \min\{\delta, \max\{-\delta, \parallel\overrightarrow{^wC^wP}\parallel-\parallel\overrightarrow{^wC^wV}\parallel\}\}
\end{equation}

For the TSDF weighting method, a constant weighting policy can be applied to the TSDF construction for simplicity. While aiming to improve the quality of details in the mapping result, a weighting strategy considering both depth value \(z\) and behind-surface voxels, which is proposed by \cite{bylow2013real, oleynikova2017voxblox}, is applied.

\vspace{-0.4cm}
\begin{equation}
  w = \left\{
  \begin{array}{ll}
    \frac{1}{z^2},                                           & -\sigma_{map} < sdf \leq \delta  \\
    \frac{1}{z^2}\frac{\delta + sdf}{\delta - \sigma_{map}}, & -\delta \leq sdf < -\sigma_{map}
  \end{array}
  \right.
\end{equation}

Now for a TSDF voxel \(v\) along the ray, its distance value \(v.sdf\) and weight \(v.w\) can be updated as

\vspace{-0.4cm}
\begin{equation}
  v.sdf = \frac{v.sdf*v.w+sdf*w}{v.w+w} \text{, } v.w = v.w+w
\end{equation}


Thanks to that TSDF construction inherently supports the reversible updating, we can de-integrate redundant or drifted frames from the mapping result to maintain global consistency by re-integration (See \ref{method: reintegration}). To de-integrate a frame from the map, \(v\) can be updated in reverse as

\vspace{-0.4cm}
\begin{equation}
  v.sdf = \frac{v.sdf*v.w-sdf*w}{v.w-w} \text{, } v.w = v.w-w
\end{equation}




One important function of the mapping framework is to provide information about known and unknown space to our exploration planner. Traditionally, the voxel is regarded as known if its TSDF weight is not zero, i.e. has been observed by at least once. To efficiently execute frontier-based exploration planning and maintain a good observation quality for mapping at the same time, we set a weight threshold \(\tau_w > 0.0\) as the criterion for known and unknown voxels recognition, where

\vspace{-0.2cm}
\begin{equation}
  v.state = \left\{
  \begin{array}{ll}
    \text{Unknown}, & 0.0 < v.w \leq \tau_w \\
    \text{Known},   & \tau_w \leq v.w
  \end{array}
  \right.
\end{equation}




By applying this strategy, we can make sure every place in the map can get enough observations by a proper \(\tau_w\) and larger \(\tau_w\) means more observations are needed to change the area to a known area in exploration planning.


\subsection{Online Re-integration}
\label{method: reintegration}
When loop closure is detected, a 4-DOF pose graph optimization is performed with multiview constraints from loop closing frames and camera poses on the pose graph are relocalized \cite{qin2018vins}, in which the global consistency of state estimation is obtained.

In order to rectify the map distortion correspondingly, all keyframes with correction on its corresponding camera pose are de-integrated from the map with its drifted pose and re-integrated with its rectified pose \cite{dai2017bundlefusion}. Keyframes are selected from input frames  in the keyframe database based on the method introduced in \ref{method: keyframe}.
The number of keyframes that are going to be re-integrated depends on the size of the subgraph rectified by loop closure. A large rectified subgraph of the rectified pose graph will generate a long list of keyframes to be re-integrated, which will cause a load burst on computing and might block the normal integration process. To reduce its influence, keyframes that need re-integration are put into a waiting list and then distributed to following regular integration cycles. Besides, for safety flight concerns, local maps have a higher demand for map consistency. Therefore, keyframes in current and neighbor blocks have utmost priority during the re-integration process. The blocks are obtained by space partitioning illustrated in \ref{method: keyframe}.



\subsection{Redundant Frame Pruning}
\label{method: keyframe}



\begin{algorithm}[t]
  \caption{Keyframe Selection}
  \label{alg: keyframealg}
  \begin{algorithmic}[1]
    \renewcommand{\algorithmicrequire}{\textbf{Input:} Point cloud frames list \(\mathcal{L}\)}
    \renewcommand{\algorithmicensure}{\textbf{Output:} Keyframe list \(\mathcal{L}_K\)}
    \REQUIRE
    \ENSURE

    \STATE{\(gain \leftarrow \infty\)}
    \WHILE{\(gain > \tau_{gain} \)}
    \FOR{\textbf{each} \(f \in \mathcal{L_{R}}\)}
    \IF{\(f\) is not selected}
    \STATE{\(f.n_{all} \leftarrow\) number of grids \(f\) covers}
    \STATE{\(f.n_{covered} \leftarrow 0\)}
    \FOR{\textbf{each} \(g \in f.coveredGridsList\)}
    \IF{\(g.visbilityCount == 0\)}
    \STATE{\(f.n_{covered} \leftarrow f.n_{covered} + 1\)}
    \ENDIF
    \ENDFOR
    \STATE{\(f.n_{new} \leftarrow f.n_{all}-f.n_{covered}\)}
    \ENDIF
    \ENDFOR
    \STATE{}
    \STATE{\(f^* \leftarrow \underset{f}{\arg \min} \frac{f.cost}{f.n_{new}}\)}
    \STATE{\(gain \leftarrow f^* .n_{new}\)}
    \STATE{\(f^*.selected \leftarrow true\)}
    \STATE{\(\mathcal{L}_K\).push(\(f^*\))}
    \STATE{}
    \FOR{\textbf{each} \(g \in f^*.coveredGridsList\)}
    \IF{\(g.visbilityCount > 0\)}
    \STATE{\(g.visbilityCount \leftarrow g.visbilityCount - 1 \)}
    \ENDIF
    \ENDFOR
    \ENDWHILE
  \end{algorithmic}
\end{algorithm}

\begin{figure}[t]
  \begin{center}
    {\includegraphics[width=0.6\columnwidth]{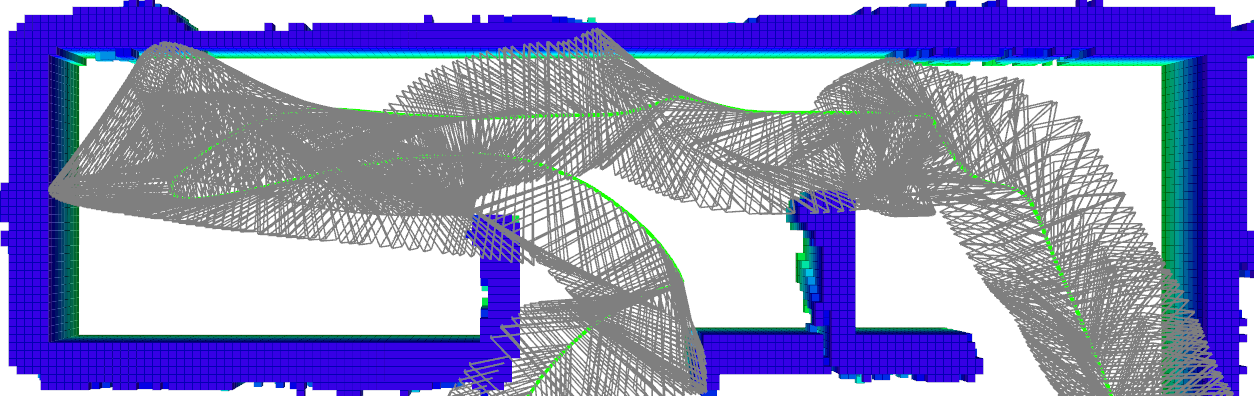}}
  \end{center}
  \begin{center}
    {\includegraphics[width=0.6\columnwidth]{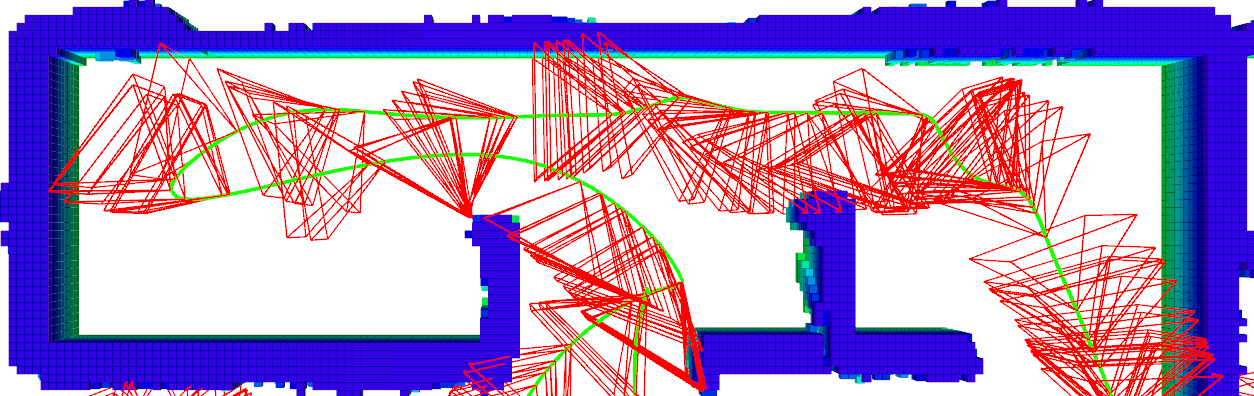}}
  \end{center}
  \vspace{-0.4cm}
  \caption{Illustration of the frame pruning. The pyramids represent the frame's FOV. In the top image, inessential frames (grey) will decelerate the re-integration process. In the bottom picture, keyframes (red) are selected from above frames by adopting set cover formulation.}
  \label{fig:set_cover2}
  \vspace{-0.4cm}
\end{figure}

The objective of the frame pruning process is to remove redundant point cloud frames that provide much duplicate information about map details but cumber the re-integration process, as shown in Fig. \ref{fig:set_cover2}. To identify the representative frames that do not contain superfluous information and enable real-time re-integration, we form this selection criterion as the well-studied un-weighted Set Cover Problem (SCP).

The universe set \(U\) is a voxel grid in the explored area with resolution \(\sigma_{scp}\). As this voxel grid only served for the checking of space observed by each frame, the resolution requirements here is lower than mapping which needs clear details about obstacles. We set \(\sigma_{scp}=2\sigma_{map}\) to accelerate visibility check for each frame without significant changes on frame's coverage. For frame \(f\), the voxels \(\{v_{f,1}, v_{f,2}, ..., v_{f,n_f}\}\) in \(U\) within its Field of View (FOV) are gathered as a subset \(S_f \subseteq U\) with cost \(c_f = 1.0\) for an un-weighted set cover problem. The cost effectiveness \(\alpha\) for frame \(f\) is defined as follows, where \(f.n_{new}\) is the number of new voxels the frame \(f\) covered.

\vspace{-0.4cm}
\begin{equation}
  \alpha_f=\frac{c_f}{f.n_{new}}
\end{equation}

\begin{figure}
  \begin{center}
    \subfigure[Actual scene]{\includegraphics[width=0.32\columnwidth]{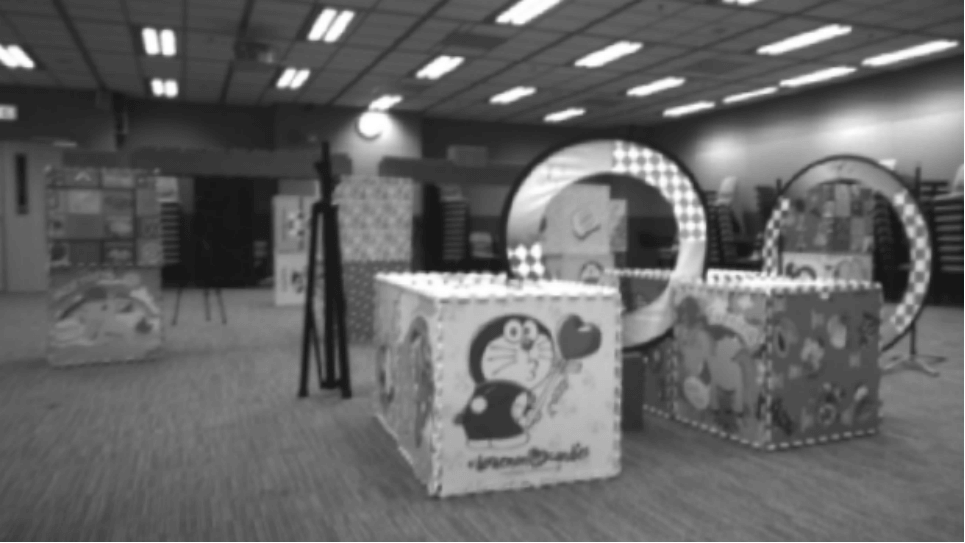}}
    \subfigure[Frame pruning with \(\tau_{gain}=0\)]{\includegraphics[width=0.32\columnwidth]{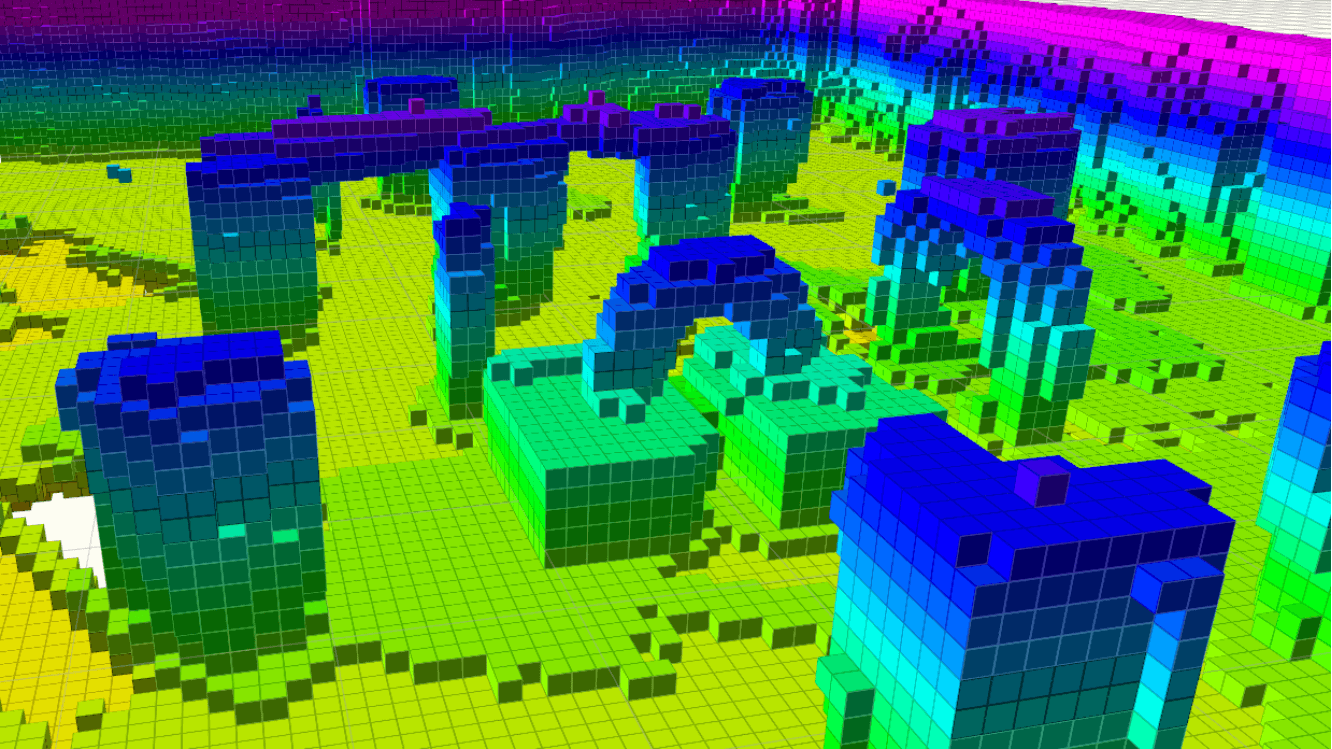}}
    \subfigure[Frame pruning with \(\tau_{gain}=50\)]{\includegraphics[width=0.32\columnwidth]{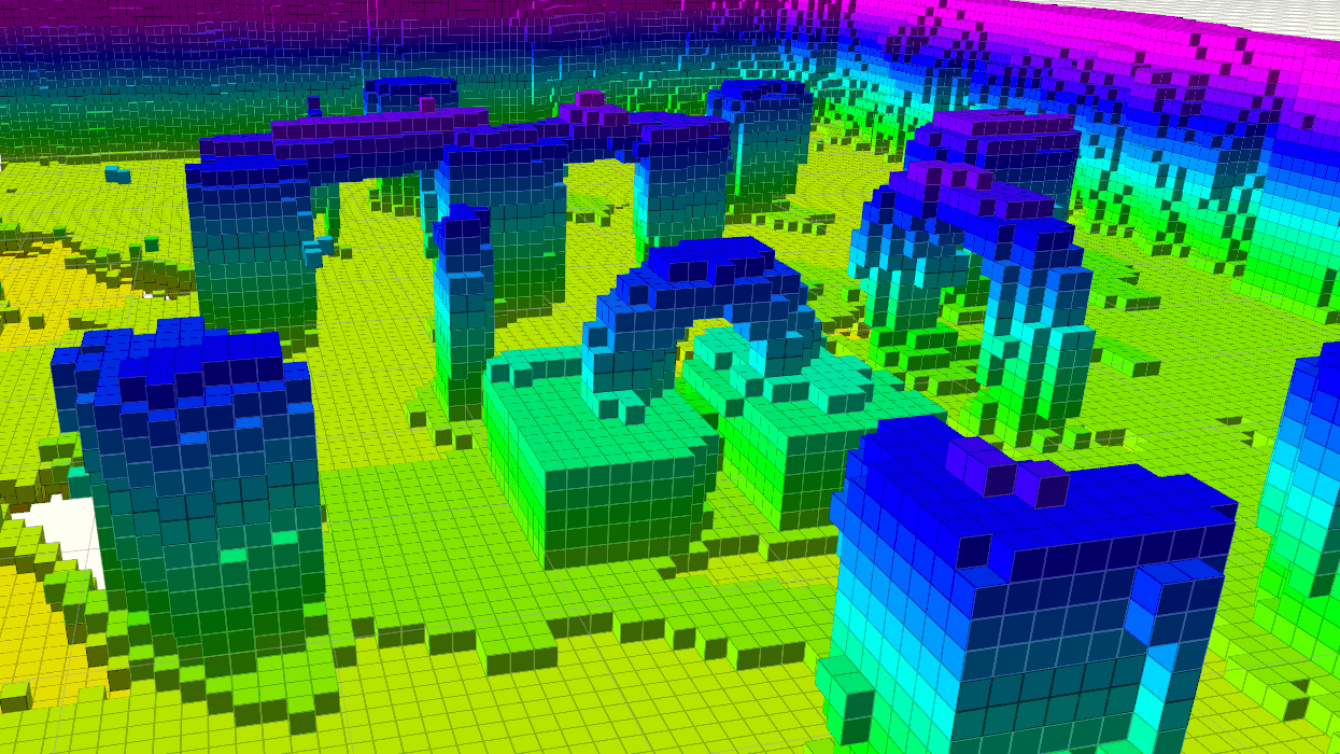}}
  \end{center}
  \vspace{-0.4cm}
  \caption{The frame pruning and a small value of \(\tau_{gain}\) has little effect on the map quality. The number of frames in (b), (c) is 544 and 456 respectively. The total number of frames is 1468.}
  \label{fig:set_cover}
  \vspace{-1.5cm}
\end{figure}

In the classic greedy algorithm for set cover problem, the goal will be formed as that for \(k\) frames, we need to find a keyframe set \(K \subseteq \{S_1, S_2, ..., S_k\}\) with minimum total cost \(c_{total} = \sum_{i \in [1,k]} c_i\), such that \(\cup_{i \in [1,k]} S_i = U\).
We make several changes based on the classic algorithm to make the implementation compatible and practical with our framework.
First, we make a minimum number of observations \(n_{ob}\) for each voxel and each voxel in \(U\) will be initialized with a visibility counter with value \(n_{ob}\). Only when the visibility counter of one voxel is deducted to zero, this voxel will be marked as covered and more observation on this voxel will be disposed. This is because covering each voxel in \(U\) just once is not enough, which will be easily disturbed by noise or drifted pose.
With this manner, the \(n_{new}\) of frame \(f\) can be calculated as the difference between total number of its visible voxels and how many voxels among these are already marked as covered.
Second, we modify the termination criterion from \(\cup_{i \in [1,k]} S_i = U\) to the gain of one cycle , which is counted as the number of new voxels covered by this frame in \(U\), is less than a certain threshold \(\tau_{gain}\).
Ideally, \(\tau_{gain}\) is set to 0 to ensure a full coverage.
However, in practice, the solution got from the classic greedy algorithm usually have a number of frames providing little new information compared with tens of thousands voxels covered by each frame, i.e. \(f.n_{new} \ll f.n_{all}\). Such a few new voxels have almost no effect on the mapping result as shown in Fig. \ref{fig:set_cover}.
Therefore, we assign \(\tau_{gain}\) with a small value (\(\sim\frac{10}{\sigma_{scp}}\)), which can reduce the keyframe amount and accelerate re-integration process without loss of quality.
With the initial visibility count for each voxel set to 1 and \(\tau_{gain}\) set to 0.0, our keyframe selection will degenerate to the classic greedy method.
More details can be found in Algorithm \ref{alg: keyframealg}.

Solving the set cover problem for the frames on the entire path is extremely time-consuming after a long flight.
We adopt space partitioning and divide the map into equal-size axis-aligned blocks and execute set cover solver only on blocks with new frames.
For \(N\) frames, this can reduce the time complexity from \(O(\log N)\) to \(O(M_{new}\log \frac{N}{M_{all}})\), where \(M_{all}\) and \(M_{new}\) are number of all blocks and blocks with new frames and \(M_{new} \ll M_{all}\) typically.




\subsection{Exploration Planning with Active Loop Closure}
\label{method:alc}

Since the mapping result relies on globally consistent localization, which requires more loop closure to be detected, we propose an exploration planning with Active Loop-Closure  (ALC) technique based on our previous approach \cite{zhou2021fuel}.
The key idea is to find an exploration tour that efficiently visits all frontier clusters frontier clusters\cite{yamauchi1997frontier} as well as potential loop-closing sites.

\begin{figure}
  \begin{center}
    {\includegraphics[width=0.7\columnwidth]{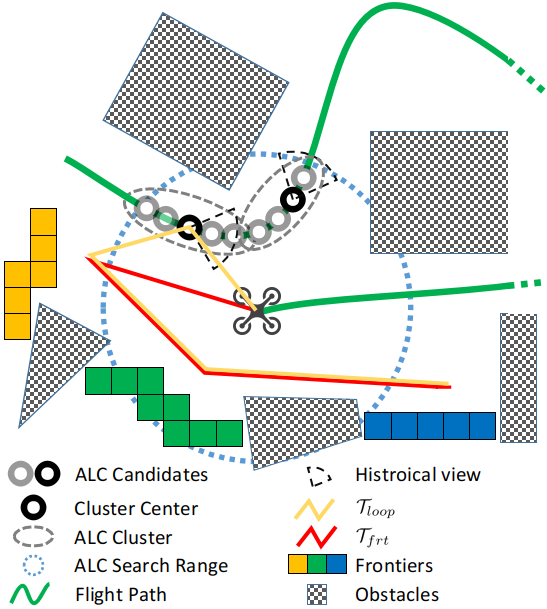}}
  \end{center}
  \vspace{-0.4cm}
  \caption{\label{fig:alc} An illustration of the proposed active loop closure method. The UAV will search loop closure candidate within a certain range and evaluate the cost of different global tours.}
  \vspace{-1.2cm}
\end{figure}

To identify the candidate location where the quadrotor may have an opportunity to detect a loop-closure, consecutive historical viewpoints along the flight path are grouped into clusters, as illustrated in Fig. \ref{fig:alc}.
We empirically constrain the size of each cluster in order to distinguish distinct regions previously visited by the quadrotor.
To efficiently find the nearby viewpoints, which are considered by our exploration planning, we also maintain a KD-Tree for the historical viewpoints\footnote{Though the viewpoint belongs to $ (x,y,z,\phi) $, the construction and query of KD-Tree here only consider the viewpoint's position.} in the flight path (Line 1, Algorithm \ref{alg: active_loop}).

Firstly, we consider the active loop closure clusters into the global tour planning \cite{zhou2021fuel}.
In \cite{zhou2021fuel}, a global tour passing only frontier clusters is generated, while in our method we also consider previously visited viewpoints.
If a loop closure cluster candidate exists nearby the quadrotor,
we choose a cluster with minimum cost from current state to its center viewpoint.
Next, an optimal tour \(\mathcal{T}_{loop}\) that visits the candidate first and all frontier clusters subsequently is found by solving an Asymmetric Traveling Salesman Problem (ATSP).
To this end, a cost matrix \(M_{loop}\) containing the cost information among all frontiers and the selected ALC cluster is computed.
The cost calculation of two viewpoints utilizes the time lower bound required to move between them proposed in \cite{zhou2021fuel}.
In order to avoid large detour which will significantly reduce the exploration efficiency, the optimal tour \(\mathcal{T}_{frt}\) considering only frontier cluster is also generated and compared with \(\mathcal{T}_{loop}\).
To find \(\mathcal{T}_{frt}\), a cost matrix \(M_{frt}\), which contains the cost information among all frontiers and current state is also generated.
Note that although an extra tour $ \mathcal{T}_{loop} $ is computed comparing with \cite{zhou2021fuel}, it does not introduce significant overhead, as $ M_{loop} $ can be obtained trivially by slightly modifying $ M_{frt} $, as shown in Line 7-10, Algorithm \ref{alg: active_loop}.
Then the next goal of flight is decided by comparison between the total cost of original global tour with only frontiers \(C_{frt}\) and the cost of global tour containing the selected ALC cluster \(C_{loop}\).

\vspace{-0.4cm}
\begin{equation}
  C_{loop}  \le (1+\epsilon) C_{frt}
\end{equation}

Since active loop closure will usually take a detour which reduce the exploration efficiency, different \(\epsilon\) denotes a tradeoff between the exploration efficiency and global consistency. The greater \(\epsilon\) means the planner will tolerate more cost spending on loop closure and maintain a more globally consistent map.

\begin{algorithm}[t]
  \caption{Active Loop Closure Planning}
  \label{alg: active_loop}
  \begin{algorithmic}[1]
    \renewcommand{\algorithmicrequire}{\textbf{Input:}}
    \renewcommand{\algorithmicensure}{\textbf{Output:}}
    \REQUIRE Flight path \(\mathcal{P}\), current position \(p\),  cost matrix \(M_{frt}\) and global tour \(\mathcal{T}_{frt}\) with only frontiers, frontier list \(FrtList\)
    \ENSURE  Global exploration tour \(\mathcal{T}_G\)
    \STATE{\(loopClusters \leftarrow nearestNeighbors(p)\) }
    \FOR{\textbf{each} \(c \in loopClusters\)}
    \STATE{\(c.cost \leftarrow computeCost(p, c)\)}
    \ENDFOR
    \STATE{\(c^* \leftarrow \underset{c}{\arg \min} \text{ } c.cost\)}
    \STATE{}
    \STATE{\(M_{loop} \leftarrow M_{frt}\)}
    \FOR{\textbf{each} \(frt \in FrtList\)}
    \STATE{\(M_{loop}(0, frt.index) \leftarrow computeCost(c^*, frt)\)}
    \ENDFOR
    \STATE{\(\mathcal{T}_{loop} \leftarrow solveAsymmetricTSP(M_{loop})\)}
    \STATE{}
    \STATE{\(C_{frt} \leftarrow 0\), \(C_{loop} \leftarrow 0\)}
    \STATE{\(computeTotalCost(p, \mathcal{T}_{frt}, M_{frt}, C_{frt})\)}
    \STATE{\(computeTotalCost(p, \mathcal{T}_{loop}, M_{loop}, C_{loop})\)}
    \STATE{\(C_{loop} \leftarrow C_{loop} + c^*.cost\)}
    \IF{\(C_{loop} < (1+\epsilon)C_{frt}\)}
    \STATE{\(\mathcal{T}_G \leftarrow \mathcal{T}_{frt}\)}
    \ELSE
    \STATE{\(\mathcal{T}_G \leftarrow \mathcal{T}_{loop}\)}
    \ENDIF
    \STATE{}
    \STATE{\textbf{function } \(computeTotalCost(p, \mathcal{T}, M, C)\):}
    \STATE\hspace{\algorithmicindent}{\(vpLast \leftarrow p \)}
    \STATE\hspace{\algorithmicindent}{\textbf{for each} \(vp \in \mathcal{T}\) \textbf{do}}
    \STATE\hspace{\algorithmicindent}\hspace{\algorithmicindent}{\(C \leftarrow C + M(vpLast, vp)\)}
    \STATE\hspace{\algorithmicindent}\hspace{\algorithmicindent}{\(vpLast \leftarrow vp\)}
    \STATE\hspace{\algorithmicindent}{\textbf{end for}}
  \end{algorithmic}
\end{algorithm}


As the global planning only decide if the active loop closure is needed and select the center viewpoint, more viewpoints in the ALC clusters besides the center viewpoint are taken into consideration for local viewpoints refinement based on the graph search-based approach in \cite{zhou2021fuel}, which further improves the path quality and allow faster exploration.
Finally, a minimum-time trajectory is generated towards the next viewpoint along the refined path through an efficient gradient-based optimization framework\cite{zhou2019robust}, enabling safe and agile navigation.
The quadrotor will be detached from local ALC clusters if no loop closure is successfully detected within a certain period, which prevents distraction from exploration the task.
Note that we may evaluate the possibility of loop-closure for historical viewpoints as \cite{lee2021real} does, which is left as a future work.

\section{Experiment Result \& Comparison}
\label{sec:evaluation}
\subsection{Benchmarked Simulation}

\begin{figure*}[t]
  \centering
  \includegraphics[width=0.19\linewidth]{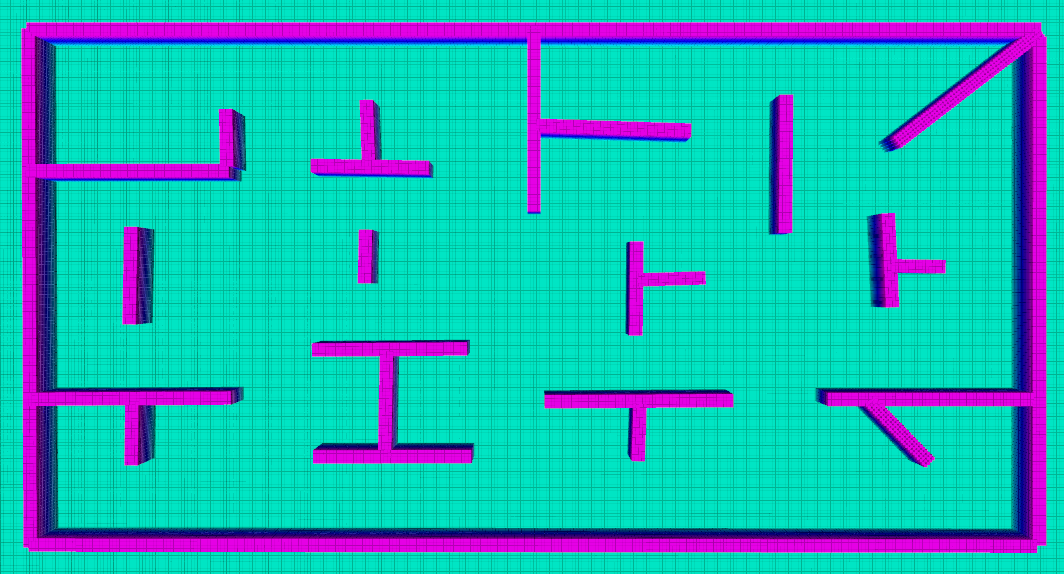}
  \includegraphics[width=0.19\linewidth]{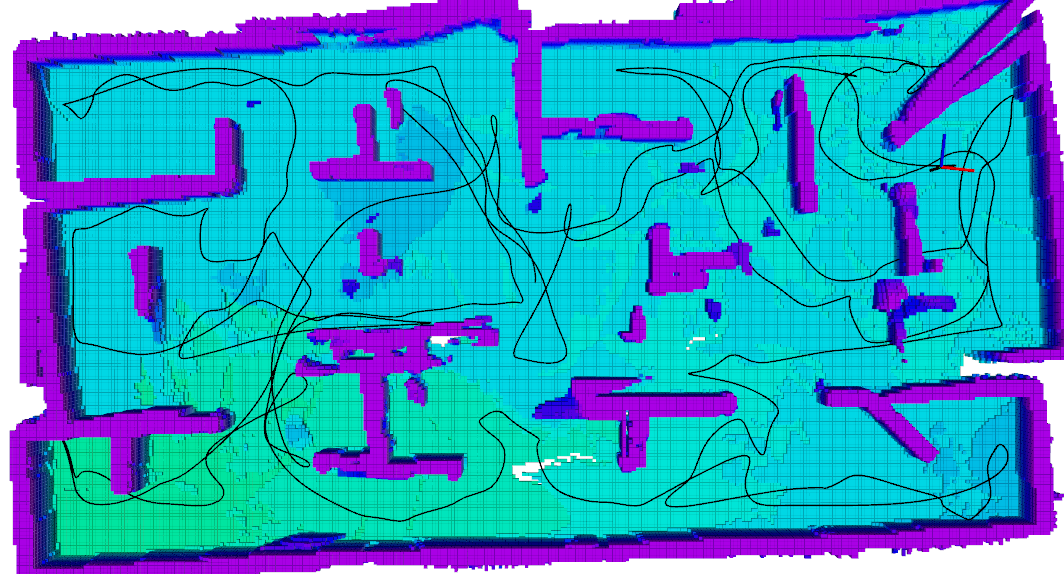}
  \includegraphics[width=0.19\linewidth]{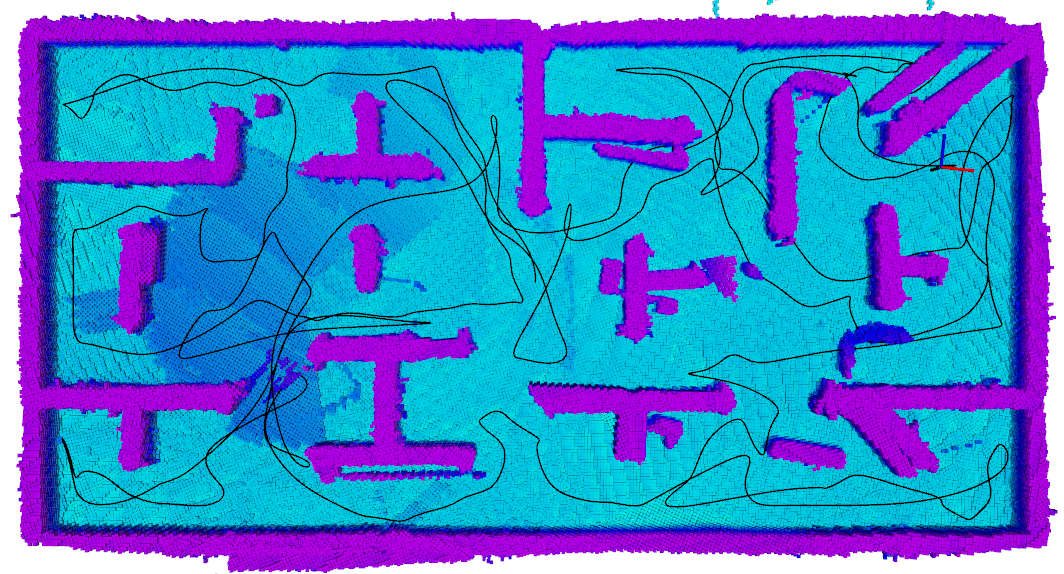}
  \includegraphics[width=0.19\linewidth]{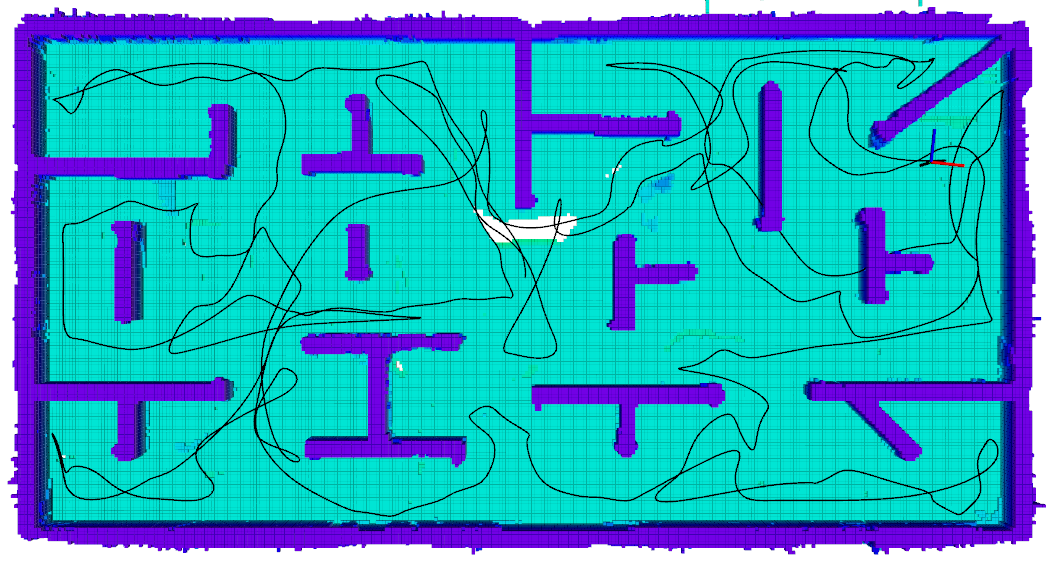}
  \includegraphics[width=0.19\linewidth]{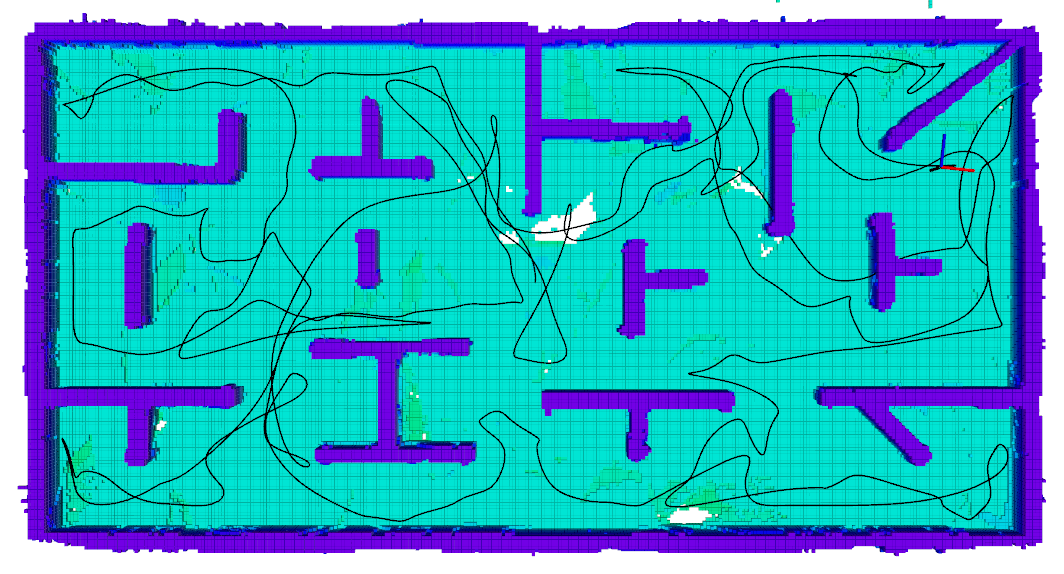}
  \caption{Mapping result of exploration in simulation environment (\(30 \times 16 \times 3\) m\(^3\)) under severe odometry drift (s4). From left to right: the ground truth point cloud, Voxblox, C-blox, our method w/o frame pruning and our method. It can be found that our method can maintain a good global consistency during the exploration flight. If the loop closure result is close to the actual localization, our reconstruction result is approximate to the ground truth. }
  \vspace{-0.4cm}
  \label{fig:sim_all}
\end{figure*}

Extensive experiments are conducted in simulation environment. We compare our mapping result obtained from autonomous exploration with other representative works. For a better evaluation on the accuracy and robustness of the methods, we set four different level of odometry drift from s1 to s4 by applying a Gaussian Noise \(v_{xyz} \sim N(\mu_{xyz},\sigma_{xyz}^2)\) and \(\omega_{\phi} \sim N(\mu_{\phi},\sigma_{\phi}^2)\) on each odometry updating cycle with frequency 200Hz. Details of the drift parameter setting can be found in Table \ref{tab:drift_param}. The dynamic parameters are set as \(v_{max} = 2.0\) m/s, \(a_{max} = 2.0\) m/s\(^2\) and \(r_{max} = 0.9\)  rad/s.
We uniformly sample a collection of zero-level iso-surface points on the TSDF. For each of the point, a nearest point from the ground truth point cloud is found to compute the distance error.

\begin{table}
  \centering
  \caption{drift parameter setting }
  \label{tab:drift_param}
  \begin{tabular}{c|c|c|c|c}
    \hline\hline
    param (m/s)       & s1  & s2   & s3   & s4     \\
    \hline
    \(\mu_{xyz} \)      & 0.0 & 0.0  & 0.0  & 0.0    \\
    \(\sigma_{xyz}^2 \) & 0.0 & 2e-2 & 5e-2 & 8e-2   \\
    \(\mu_{\phi}\)      & 0.0 & 0.0  & 1e-3 & 1.5e-3 \\
    \(\sigma_{\phi}^2\) & 0.0 & 2e-2 & 5e-2 & 8e-2   \\
    \hline\hline
  \end{tabular}
\end{table}

As the result shown in Fig. \ref{fig:sim_all} and Table \ref{tab:result_sim}, four methods have a similar performance under negligible odometry drift. When it comes to  odometry drift, the result of Voxblox\cite{oleynikova2017voxblox} distorts the most as 
it does not include rectifying global map according to input loop closure results. C-blox\cite{millane2018c} can support such map rectification by correct each submap with rectified camera localization. 
However, it still has a higher RMSE mainly because it assumes state estimation is consistent within a submap, which is not reasonable under large drift. The result of our method w/o frame pruning, i.e. re-integrate all frames, shows a similar result as ours, but its real-time performance is poor on onboard computer.

\begin{figure}[t]
  \begin{center}
    {
      \includegraphics[width=0.32\columnwidth]{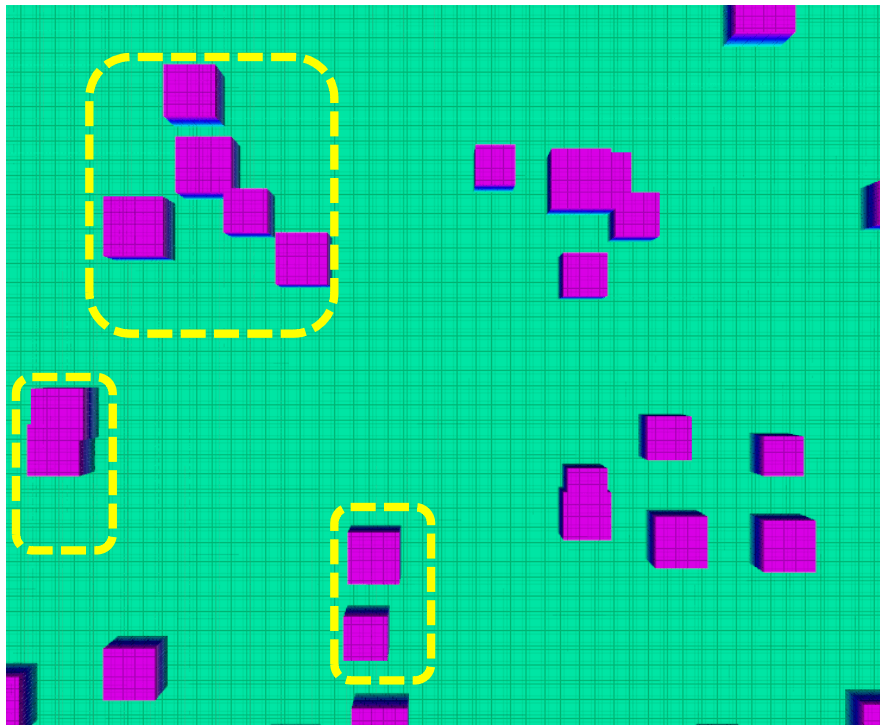}
      \includegraphics[width=0.32\columnwidth]{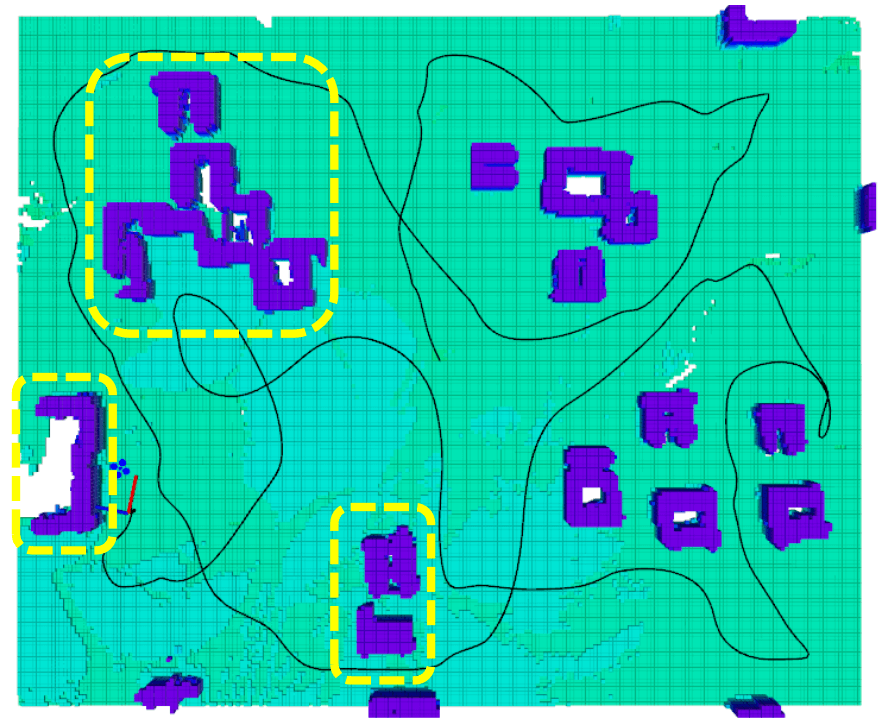}
      \includegraphics[width=0.32\columnwidth]{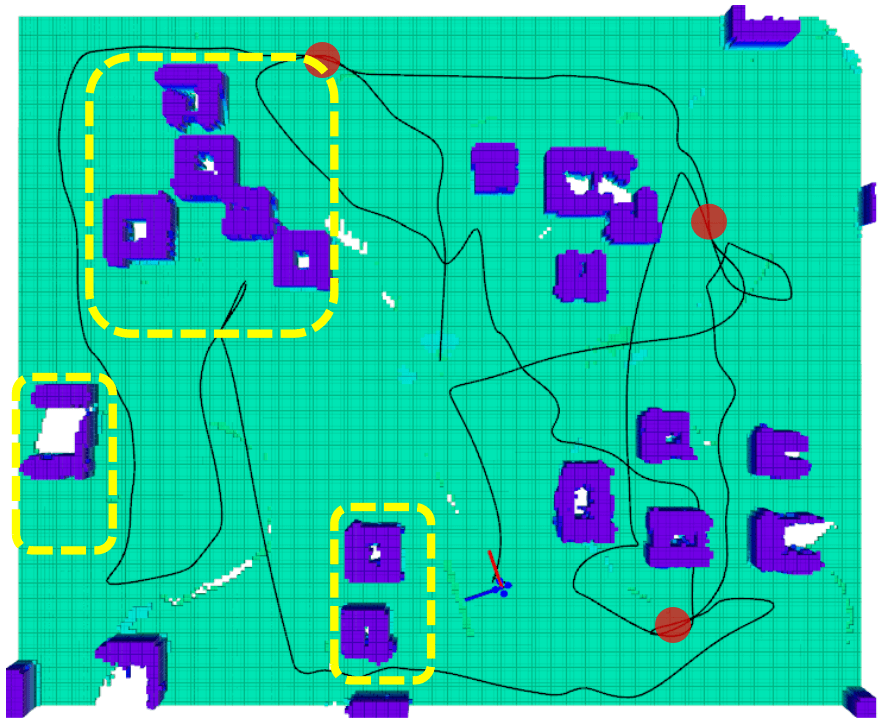}
    }
  \end{center}
  \vspace{-0.4cm}
  \caption{Left: Ground truth (\(20 \times 16 \times 3\) m\(^3\)), Middle: w/o ALC; Right: w/ ALC. Successful loop closure is marked as red dot.}
  \label{fig:time_alc}
  \vspace{-0.4cm}
\end{figure}

We also compare the time cost for exploration w/ and w/o active loop closure under s3 drift in Table \ref{tab:drift_param}, which is 115s(w/o ALC) and 124s(w/ALC). The exploration flight is shown in Fig. \ref{fig:time_alc}. With \(\epsilon = 0.2\) in Sect. \ref{method:alc}, the time of exploration planning with ALC is only 8\% more than that w/o ALC but it gains 3 more opportunity to detect loop closure. The final mapping result shows that exploration w/ ALC has superior performance under odometry drift.

\begin{table}
  \centering
  \caption{\label{tab:benchmark} Accuracy comparison }
  \label{tab:result_sim}
  \begin{tabular}{ccccc}
    \hline\hline
    \multirow{2}{*}{\textbf{Method}}    & \multicolumn{4}{c}{\textbf{RMSE (m)}}                                                    \\
    \cline{2-5}
    \multicolumn{1}{c}{}                & \textbf{s1}                           & \textbf{s2}    & \textbf{s3}    & \textbf{s4}    \\
    \hline
    Voxblox\cite{oleynikova2017voxblox} & 0.092                                 & 0.084          & 0.169          & 0.266          \\
    C-blox\cite{millane2018c}           & 0.109                                 & \textbf{0.081} & 0.147          & 0.157          \\
    Ours(w/o frame pruning)                & \textbf{0.089}                        & 0.091          & 0.098          & \textbf{0.085} \\
    Ours                                & \textbf{0.089}                        & 0.087          & \textbf{0.094} & 0.087          \\
    \hline\hline
  \end{tabular}
  \vspace{-2.3cm}
\end{table}



\subsection{Field Test}

\begin{figure}[t]
  \begin{center}
    {
      \includegraphics[width=0.32\columnwidth]{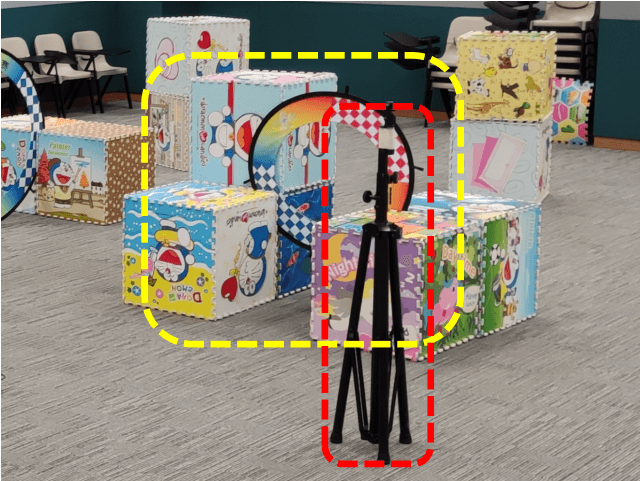}
      \includegraphics[width=0.32\columnwidth]{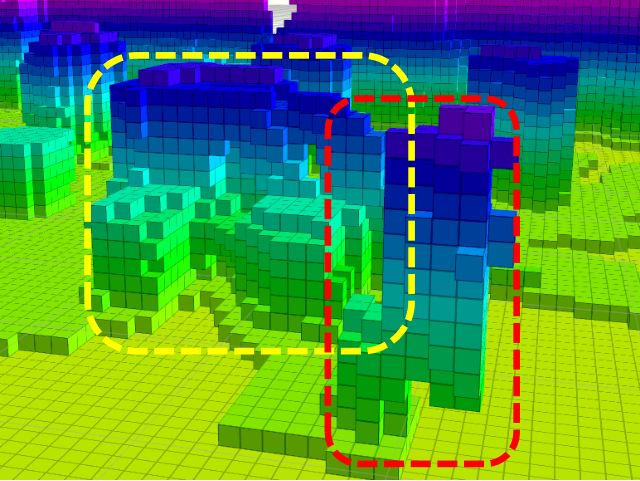}
      \includegraphics[width=0.32\columnwidth]{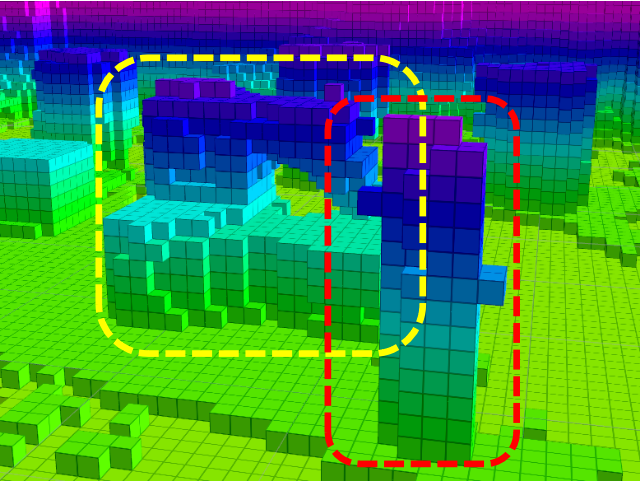}
    }
  \end{center}
  \begin{center}
    {
      \includegraphics[width=0.32\columnwidth]{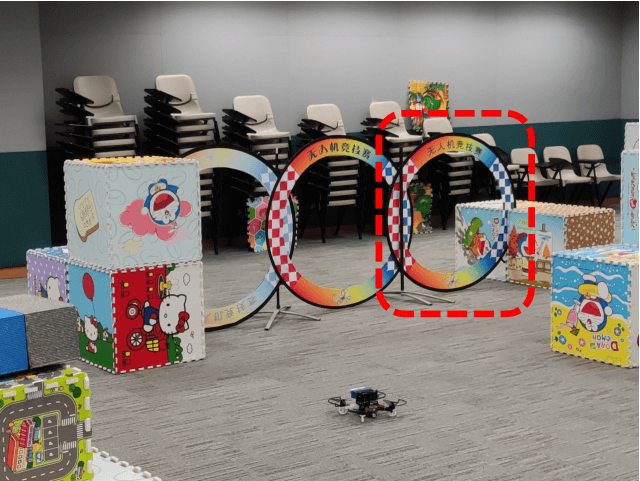}
      \includegraphics[width=0.32\columnwidth]{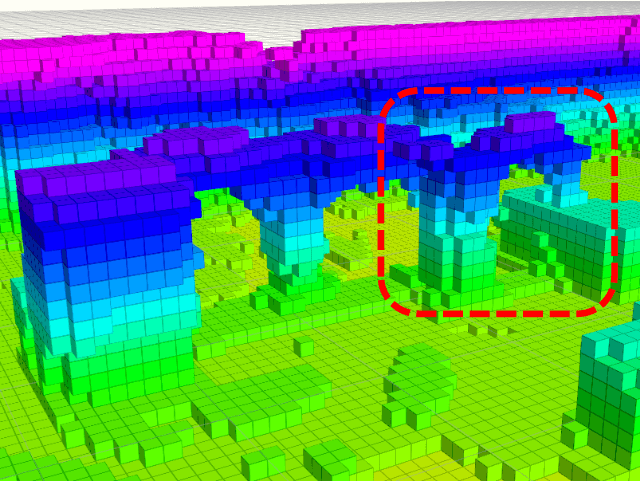}
      \includegraphics[width=0.32\columnwidth]{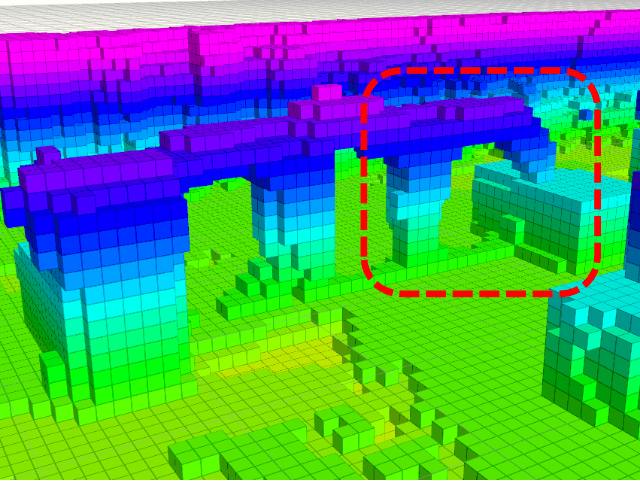}
    }
  \end{center}
  \vspace{-0.4cm}
  \caption{From left to right: the intercepted actual scene from Fig. \ref{fig:intro}, the reconstruction result from FUEL\cite{zhou2021fuel} and ours. For the first row, the result from \cite{zhou2021fuel} has strong distortion on the ring and the tripod. On the second row, our method reconstructs three coplanar rings well.}
  \label{fig:exp_result}
  \vspace{-0.8cm}
\end{figure}

In order to further evaluate the performance of our proposed approach, comparative field experiments are conducted in a large classroom with cluttered environment between the FUEL\cite{zhou2021fuel} and ours.
The bounded size of the exploration area is \(20 \times 14 \times 3\) m\(^3\).
With the installation of a RealSense depth camera D435i and a DJI Manifold 2-C onboard computer powered by Intel Core i7-8550U, the quadrotor platform fully relies on onboard sensors and computation to perform the exploration task without any external devices.
The onboard state estimation and loop closure method is VINS-Fusion\cite{qin2018vins}.
The linear velocity, acceleration and yaw rate limits of both methods are set to \(v_{max} = 1.0\) m/s, \(a_{max} = 0.5\) m/s\(^2\) and \(r_{max} = 0.6\)  rad/s.

A global map and local details comparison between FUEL and ours is shown in Fig. \ref{fig:intro} and \ref{fig:exp_result}. Though loop-closure of localization is enabled for both methods, the mapping result from FUEL still has several distortions. In our result, a globally consistent map is maintained.
The flight path in Fig. \ref{fig:intro} labeled with black line shows the effect from the active loop closure planning.
This experiment demonstrates that our novel approach has the ability to maintain a global consistent mapping result during autonomous exploration.


\section{Conclusions}
\label{sec:conclude}

In this paper, we propose a systematic mapping and planning framework for exploration under odometry drift, which can maintain the global consistency of the exploration map. The mapping is based on an online re-integration method with frame-pruning to enable real-time performance. Moreover, we consider the possible loop closure viewpoints actively into the exploration planning to improve the mapping result. The proposed method is evaluated by extensive experiments in both benchmarked simulation and real-world environment. The results show the effectiveness of our proposed method.


\addtolength{\textheight}{0.cm}

\newlength{\bibitemsep}\setlength{\bibitemsep}{0.0\baselineskip}
\newlength{\bibparskip}\setlength{\bibparskip}{0.1pt}
\let\oldthebibliography\thebibliography
\renewcommand\thebibliography[1]{%
  \oldthebibliography{#1}%
  \setlength{\parskip}{\bibitemsep}%
  \setlength{\itemsep}{\bibparskip}%
}

\normalem
\bibliography{zhang2022icra}

\end{document}